# Channel Model for End-to-End Learning of Communications Systems: A Survey

Ijaz Ahmad[1] and Seokjoo Shin[*]

Department of Computer Engineering,

Chosun University, Gwangju, 61452 South Korea

[1]ahmadijaz@chosun.kr, *Corresponding author: sjshin@chosun.ac.kr

*Abstract*—The traditional communication model based on chain of multiple independent processing blocks is constraint to efficiency and introduces artificial barriers. Thus, each individually optimized block does not guarantee end-to-end performance of the system. Recently, end-to-end learning of communications systems through machine learning (ML) have been proposed to optimize the system metrics jointly over all components. These methods show performance improvements but has a limitation that it requires a differentiable channel model. In this study, we have summarized the existing approaches that alleviates this problem. We believe that this study will provide better understanding of the topic and an insight into future research in this field.

## I. Introduction

The traditional communication systems are represented with a chain of multiple independent processing blocks; each executing a well-defined and isolated sub-function, e.g., source coding, modulation, channel estimation, equalization as shown in Fig. 1. [1], [2]. The block communication systems has the advantage that each block can be individually analyzed and optimized, leading to the efficient and stable systems that are available today [2]. These conventional systems has two limitations. First, such systems are known to be sub-optimal [3] and, each individually optimized blocks does not guarantee end-to-end performance of the communication systems [1]. Second, the algorithms for each block have solid foundations in statistics and information theory. These algorithms are often optimized for mathematically convenient models, which are stationary, linear and have Gaussian statistics. However, real systems have many imperfections and non-linearities, e.g. hardware imperfection and noisy channels [1].

Machine learning (ML) based communication systems deals with the fundamental problem of reproducing a sender's message exactly or approximately at the receiver's end propagated over a channel [2]. ML implements communication systems using neural networks (NN). It has the following advantages over conventional communication system: (i) ML-based communication system does not require a mathematical model for representation and transformation of information. (ii) Unlike conventional block structure, a ML-based communication system can be designed to optimize end-to-end performance of the system. (iii) Since the execution of NNs are parallel on concurrent architectures; therefore, ML algorithms can provide faster processing speed at lower energy cost than conventional systems. These advantages have motivated researchers to extend ML applications to communications and signal processing. The applications of ML in communications include belief propagation algorithm to improve channel decoding [4], [5], learning of cryptography schemes [6], one-shot channel decoding [7], as well as compression [8]. One of the interesting applications is learning an end-to-end communications system using autoencoder proposed first in [1]. The goal of the system is to optimize the end-to-end recovery accuracy of communication system. The major limitation of end-to-end paradigm is that channel state information (CSI) must be known during training time to minimize the reconstruction error rate [2], [9], [10]. Several works have been done in this direction to circumvent the missing channel gradient problem [9]–[14]. In this study, we investigate these methods and show their performance efficiency.

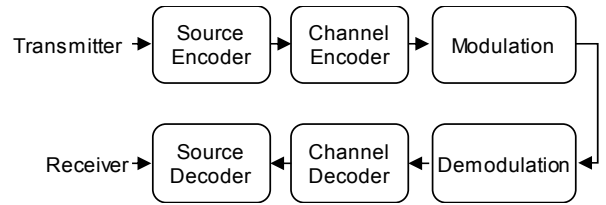

Fig. 1. Conventional block communication system model.

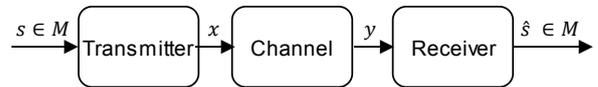

Fig. 2. Communication system with **s** as an input.

The rest of the paper is summarized as; in Section II, we present the background of ML based end-to-end communication system and autoencoder architecture. In Section III, we carried the analysis of solutions proposed to mitigate the missing channel gradient problem. Finally, conclusion of the study.

## II. ML Based End-to-end Communication Systems

A communication system consists of two nodes: transmitter and receiver aim to exchange information over a channel as shown in Fig. 2. At the transmitter, let **s** denotes the data symbols chosen from **M** possible messages. The transmitter maps **s** to representation **x**, which is to be sent over the channel to the receiver. The channel is a stochastic system whose output is **y**. The receiver task is to detect **s** from the received signal **y**. The challenge is to reconstruct the exact or approximate **s** denoted by **ŝ** from the corrupted **x** (i.e. the channel output **y**) at the receiver's end under adverse channel conditions. The corruption in received message is due to channel impairments and hardware imperfection.

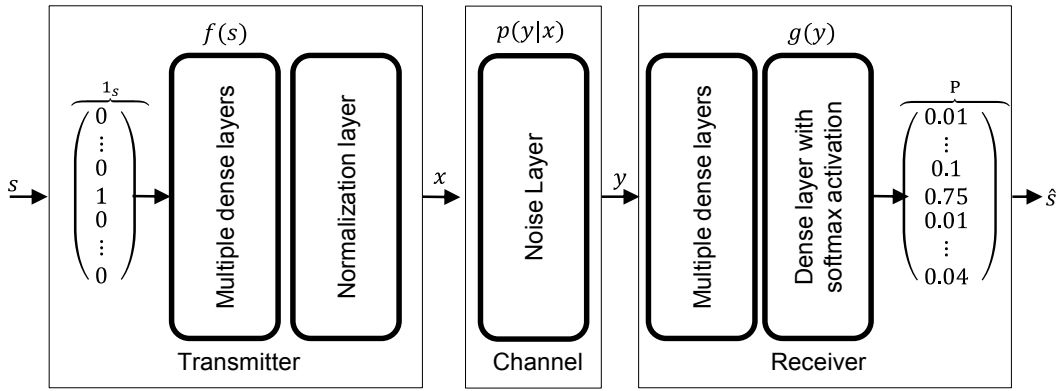

Fig. 3. Machine learning based communication system represented as an autoencoder.

In the conventional communication system design, the transmitter and receiver are divided into blocks each performing a sub-task, e.g., source coding, channel estimation, equalization and modulation. However, it is not clear that this subdivision achieves overall best possible performance. Therefore, a machine learning (ML) based communication system can be designed to optimize end-to-end performance of the system without the need of a mathematical model for representation and transformation of information.

As proposed in [1], from ML point of view, a communications system appears to be a type of autoencoder. An autoencoder learns a compressed representation of its input and allows reconstruction with a minimal error at the output [15]. The autoencoder for communication system consists of transmitter, receiver and channel, that learns **x** representation of **s**, which is robust to channel impairments. The goal is to recover the transmitted message with small error possible.

### A. Architecture of Autoencoder based Communication System

An example of autoencoder based communication system is shown in Fig. 3. The system includes the functionality of both transmitter and receiver. An input to the transmitter is the message **s**, encoded as a one-hot vector. The one-hot vector represents the conventional data in $M$-dimensional vector having only one non-zero element. Since, the data rate for the same channel environment is proportional to the number of bit being conveyed. Therefore, the one-hot vector representation limits the data transmission rate to $M$ bits. An alternative generalized data representation scheme is proposed in [16]. In this scheme, instead of using the conventional one-hot vector representation, the authors proposed a bit vector containing $m$ non-zero entries to improve the data rate.

The transmitter consists of feedforward neural network (NN) with multiple dense layers including a rectifier linear unit (ReLU) and a linear layer. To ensure that all physical constraints are met on **x**, the NN is followed by a normalization layer. The transmission channel is implemented as zero-mean additive white Gaussian noise (AWGN) layer with a fixed variance. Note that the autoencoder is suitable for any type of channel, without a tractable mathematical model, as long as real datasets are available [16]. Similar to transmitter, the receiver consists of another feedforward NN with multiple dense layers. The last layer of the receiver has a softmax activation function. The output of softmax function is a probability distribution over all messages, i.e., $p \in (0,1)^M$. The estimated message $\hat{s}$ is obtained from the index of the element with the highest probability in **p**. The autoencoder is trained in a supervised learning manner using Stochastic Gradient Descent (SGD) with large training dataset to optimize the end-to-end recovery accuracy. The dataset consists of all possible messages **s**. The most common loss function used are categorical cross-entropy and MSE. Once the NN is trained then it can be applied to practical scenarios.

### III. PERFORMANCE EVALUATION

In this section, we provide the theoretical analysis of the methods that have been proposed to mitigate the missing channel gradient problem.

The major limitation of end-to-end communication system proposed in [1] is that channel state information (CSI) must be known during training time to minimize the reconstruction error rate [2], [9], [10]. However, in real communication systems, it is hard to know CSI beforehand because the channel impairments are sometimes unknown and hard to be expressed analytically. Since, the NN uses SGD with backpropagation of error gradients; however, the missing channel transfer function blocks the backpropagation. The channel model can be assumed; e.g. additive white Gaussian noise (AWGN) as in [1]. However, the assumption would bias the learned weights [9]. One solution to mitigate this problem is a two-phase training strategy proposed in [2]. The first phase is to train the autoencoder using stochastic channel model. In the second phase, the receiver is trained to communicate over an actual wireless channel. However, such an approach leads to sub-optimal system, i.e., only the decoder is optimized [11]. In [11], [12] an model-free training approach has been proposed that uses reinforcement learning based training for the transmitter and supervised training of the receiver. During transmitter training, the channel and the receiver are considered as the environment. However, to achieve competitive performance of existing methods, some prior information of the channel is required [9].

The channel random behavior due to device response, interference effects, and noise effects is difficult to model with simplified analytic models. Since the channel model is a stochastic function, the authors in [10] proposed to represent it as a conditional probability distribution. To capture the stochastic behavior, they have used variational Generative adversarial network (GAN). Unlike previous works, instead of minimizing the MSE of real channel and approximated channel model, they minimized the error between conditional probability distributions and the real channel behavior measurement. Somewhat similar approach has been adopted in [9], [13] to train a surrogate channel network using GAN. However, in [9], [10], [13] the probability distribution is still an approximation and can be improved by training GAN with more samples which leads to more complexity [14]. In [14] the authors proposed a single stage training approach to train the system without assumptions about the channel model. They have used stochastic approximation technique to calculate gradients. Unlike [10], the training and deployment of the system has been performed on an actual channel.

The existing solutions for missing channel gradient problem have their own advantages and disadvantages. In practice it is hard to obtain the real channel model. One solution is to use two phase training techniques. They have the advantage of simplicity as they do not require large samples. However, the resultant system is sub-optimal. Another solution is to use GANs to model the channel as conditional probabilities. The main drawback of this approach is the use of more samples to improve the approximation.

## IV. CONCLUSION

This study reviews the recent literature on the application of ML methods in communication systems to replace the conventional block structure with an end-to-end learning communication system represented as an autoencoder. The end-to-end paradigm based on ML is an interesting concept. However, several considerations are required when choosing a channel model for an autoencoder based communication systems. We have investigated the solutions proposed for missing channel gradient problem and summarized their strengths and weaknesses.

## ACKNOWLEDGMENT

This research is supported by Basic Science Research Program through the National Research Foundation of Korea (NRF) funded by the Ministry of Education (NRF-2018R1D1A1B07048338).

## REFERENCES

[1] T. J. O'Shea and J. Hoydis, "An Introduction to Deep Learning for the Physical Layer," *ArXiv170200832 Cs Math*, Jul. 2017, Accessed: Apr. 27, 2020. [Online]. Available: http://arxiv.org/abs/1702.00832.

[2] S. Dorner, S. Cammerer, J. Hoydis, and S. ten Brink, "Deep Learning Based Communication Over the Air," *IEEE J. Sel. Top. Signal Process.*, vol. 12, no. 1, pp. 132–143, Feb. 2018, doi: 10.1109/JSTSP.2017.2784180.

[3] E. Zehavi, "8-PSK trellis codes for a Rayleigh channel," *IEEE Trans. Commun.*, vol. 40, no. 5, pp. 873–884, May 1992, doi: 10.1109/26.141453.

[4] E. Nachmani, Y. Beery, and D. Burshtein, "Learning to Decode Linear Codes Using Deep Learning," *ArXiv160704793 Cs Math*, Sep. 2016, Accessed: May 07, 2020. [Online]. Available: http://arxiv.org/abs/1607.04793.

[5] E. Nachmani, E. Marciano, D. Burshtein, and Y. Be'ery, "RNN Decoding of Linear Block Codes," *ArXiv170207560 Cs Math*, Feb. 2017, Accessed: May 07, 2020. [Online]. Available: http://arxiv.org/abs/1702.07560.

[6] M. Abadi and D. G. Andersen, "Learning to Protect Communications with Adversarial Neural Cryptography," *ArXiv161006918 Cs*, Oct. 2016, Accessed: May 07, 2020. [Online]. Available: http://arxiv.org/abs/1610.06918.

[7] T. Gruber, S. Cammerer, J. Hoydis, and S. ten Brink, "On Deep Learning-Based Channel Decoding," *ArXiv170107738 Cs Math*, Jan. 2017, Accessed: May 07, 2020. [Online]. Available: http://arxiv.org/abs/1701.07738.

[8] T. J. O'Shea, J. Corgan, and T. C. Clancy, "Unsupervised Representation Learning of Structured Radio Communication Signals," *ArXiv160407078 Cs*, Apr. 2016, Accessed: May 07, 2020. [Online]. Available: http://arxiv.org/abs/1604.07078.

[9] H. Ye, G. Y. Li, B.-H. F. Juang, and K. Sivanesan, "Channel Agnostic End-to-End Learning Based Communication Systems with Conditional GAN," in *2018 IEEE Globecom Workshops (GC Wkshps)*, Abu Dhabi, United Arab Emirates, Dec. 2018, pp. 1–5, doi: 10.1109/GLOCOMW.2018.8644250.

[10] T. J. O'Shea, T. Roy, and N. West, "Approximating the Void: Learning Stochastic Channel Models from Observation with Variational Generative Adversarial Networks," *ArXiv180506350 Cs Eess Stat*, Aug. 2018, Accessed: May 07, 2020. [Online]. Available: http://arxiv.org/abs/1805.06350.

[11] F. A. Aoudia and J. Hoydis, "End-to-End Learning of Communications Systems Without a Channel Model," in *2018 52nd Asilomar Conference on Signals, Systems, and Computers*, Pacific Grove, CA, USA, Oct. 2018, pp. 298–303, doi: 10.1109/ACSSC.2018.8645416.

[12] F. A. Aoudia and J. Hoydis, "Model-Free Training of End-to-End Communication Systems," *IEEE J. Sel. Areas Commun.*, vol. 37, no. 11, pp. 2503–2516, Nov. 2019, doi: 10.1109/JSAC.2019.2933891.

[13] H. Ye, L. Liang, G. Y. Li, and B.-H. F. Juang, "Deep Learning based End-to-End Wireless Communication Systems with Conditional GAN as Unknown Channel," *ArXiv190302551 Cs Math*, Mar. 2019, Accessed: May 07, 2020. [Online]. Available: http://arxiv.org/abs/1903.02551.

[14] V. Raj and S. Kalyani, "Backpropagating Through the Air: Deep Learning at Physical Layer Without Channel Models," *IEEE Commun. Lett.*, vol. 22, no. 11, pp. 2278–2281, Nov. 2018, doi: 10.1109/LCOMM.2018.2868103.

[15] I. Goodfellow, Y. Bengio, and A. Courville, *Deep learning*. Cambridge, Massachusetts: The MIT Press, 2016.

[16] X. Chen, J. Cheng, Z. Zhang, L. Wu, J. Dang, and J. Wang, "Data-Rate Driven Transmission Strategies for Deep Learning-Based Communication Systems," *IEEE Trans. Commun.*, vol. 68, no. 4, pp. 2129–2142, Apr. 2020, doi: 10.1109/TCOMM.2020.2968314.